\documentclass[10pt,twocolumn,letterpaper]{article}

\usepackage{cvpr}
\usepackage{times}
\usepackage{epsfig}
\usepackage[nolist,nohyperlinks]{acronym}
\usepackage{amsmath,amssymb,amsfonts}
\usepackage{algorithmic}
\usepackage{subcaption}
\usepackage{graphicx}
\usepackage{textcomp}
\usepackage{xcolor}
\usepackage{enumitem}
\setlist[description]{itemsep=1pt}
\usepackage{booktabs}
\usepackage{makecell}
\usepackage{multirow}
\usepackage{lipsum}
\usepackage{tcolorbox}
\usepackage{pgf}
\usepackage{adjustbox}
\usepackage[numbers]{natbib}
\def\BibTeX{{\rm B\kern-.05em{\sc i\kern-.025em b}\kern-.08em
    T\kern-.1667em\lower.7ex\hbox{E}\kern-.125emX}}
\usepackage[hidelinks]{hyperref}
\begin{document}

\title{Closing the Performance Gap in Biometric Cryptosystems: \\A Deeper Analysis on Unlinkable Fuzzy Vaults}
\author{
Hans Geißner and Christian Rathgeb\\
{da/sec -- Biometrics and Security Research Group}\\
{ Hochschule Darmstadt, Germany}\\
{\tt\small hans.geissner@h-da.de, christian.rathgeb@h-da.de}
}

\maketitle
\thispagestyle{empty}
\begin{acronym}[UML]
  \acro{DET}{Detection Error Tradeoff}
  \acro{BCS}{Biometric Cryptosystem}
  \acrodefplural{BCS}{Biometric Cryptosystems}
  \acro{BTP}{Biometric Template Protection}
  \acro{EER}{Equal Error Rate}
  \acro{FMR}{False Match Rate}
  \acro{FNMR}{False Non-Match Rate}
  \acro{FRGC}{Face Recognition Grand Challenge}
  \acro{GMR}{Genuine Match Rate}
  \acro{FAS}{False Accept Security}
  \acro{CNN}{Convolutional neural network}
  \acro{ROC}{Receiver Operating Characteristic Curve}
  \acro{XOR}{Exclusive-OR}
  \acro{DBR}{Direct Binary Representation}
  \acro{LSSC}{Linearly Seperable Subcodes}
\end{acronym}
\begin{abstract}
This paper analyses and addresses the performance gap in the fuzzy vault-based \ac{BCS}. We identify unstable error correction capabilities, which are caused by variable feature set sizes and their influence on similarity thresholds, as a key source of performance degradation. This issue is further compounded by information loss introduced through feature type transformations. To address both problems, we propose a novel feature quantization method based on \it{equal frequent intervals}. This method guarantees fixed feature set sizes and supports training-free adaptation to any number of intervals.
The proposed approach significantly reduces the performance gap introduced by template protection. Additionally, it integrates seamlessly with existing systems to minimize the negative effects of feature transformation. Experiments on state-of-the-art face, fingerprint, and iris recognition systems confirm that only minimal performance degradation remains, demonstrating the effectiveness of the method across major biometric modalities.
\end{abstract}

 \section{Introduction}\label{sec:intro}
Biometric recognition, the automated recognition of individuals based on behavioural or physiological characteristics, raises critical security and privacy concerns. Unlike replaceable credentials (e.g., passwords), biometric traits (e.g., faces, voices, or fingerprints) are inherently sensitive, as they may contain information about attributes like gender or ethnicity. Further, their immutability implies that compromised biometric data (templates) permanently expose an individual’s identity. Consequently, biometric data demands stringent protection. Ideally, stored biometric templates do not leak sensitive information while enabling efficient and accurate comparisons. To achieve this, \ac{BTP} methods, which secure biometric data while enabling efficient and effective comparison, can be employed. These methods must satisfy fundamental requirements that balance privacy preservation with recognition accuracy \cite{ISO-IEC-24745-TemplateProtection-2022}. Key requirements include:
\begin{description}[align=left]
    \item[Irreversibility:] It should be infeasible to reconstruct the original biometric data from the protected template.
    \item[Unlinkability:] It should be infeasible to determine if two protected template correspond to the same individual.
    \item[Revocability and Renewability:] It should be possible to issue a new protected template without revealing additional information.
    \item[Performance Preservation:] The recognition accuracy of the protected system should be comparable to that of unprotected systems.
\end{description}
\begin{figure}
    \centering
    \includegraphics[width=\linewidth]{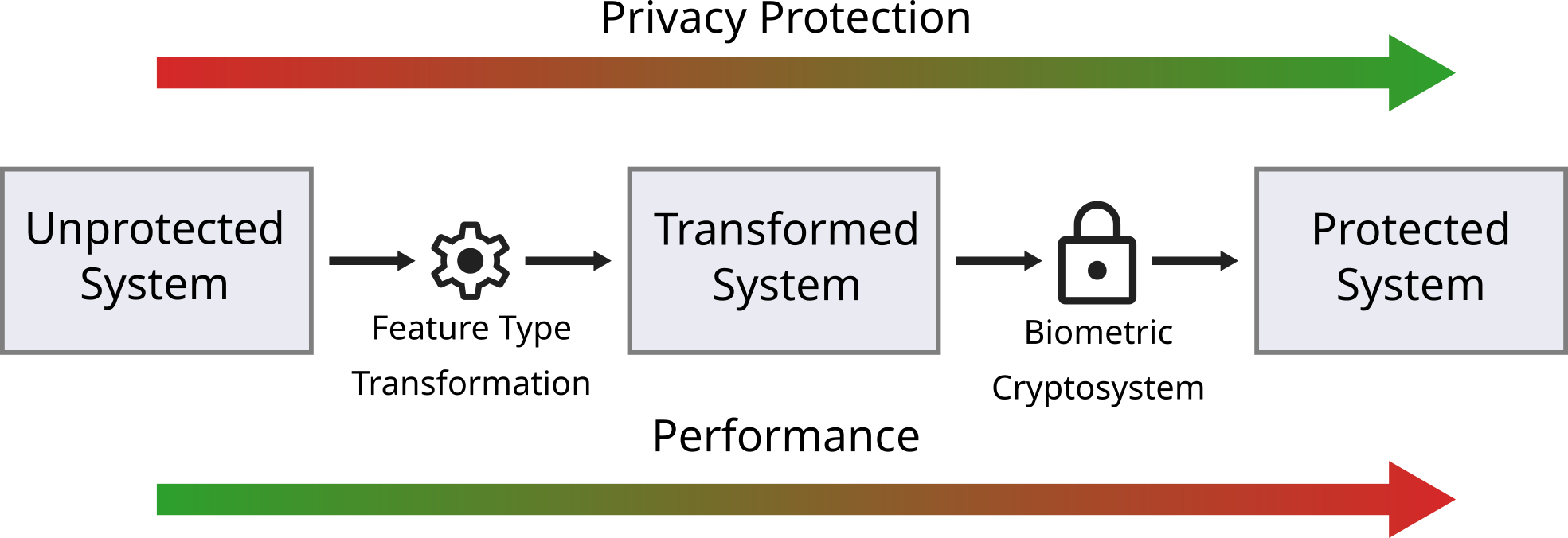}
    \caption{Impact of the application of feature-type transformations and \acp{BCS} on privacy protection and biometric performance.}
    \label{fig:performance-loss}
    \vskip -1em
\end{figure}

\ac{BCS} are a class of \ac{BTP} schemes designed to fulfill the above listed requirements.
\acp{BCS} enable the retrieval of keys using biometric data through error tolerant key generation or key binding processes that protect the original biometric data. This means that, in biometric cryptosystems, a biometric comparison is indirectly conducted by verifying the correctness of a retrieved key \cite{Uludag-BiometricCryptosystems-IEEE-2004}. Since \acp{BCS} require biometric templates in specific feature representation, so-called feature-type transformations are necessary to convert  biometric data into compatible representations. At the same time, feature-type transformations should not impair the biometric performance. Although \acp{BCS} enhance privacy protection, they are reported to reduce recognition performance \cite{NandakumarJain-BiometricTemplateProtection-SPM-2015}. \autoref{fig:performance-loss} illustrates this security-performance trade-off across three system configurations:
\begin{description}[align=left]
\item[Unprotected System:] Utilizes original templates and comparators, achieving high recognition performance but lacks privacy protection.
\item[Transformed System:] Utilizes transformed templates, creating a performance gap due to information loss during transformation.
\item[Protected System:] Integrates \ac{BCS} with transformed templates, offering privacy protection at the cost of additional performance loss relative to both unprotected and transformed systems.
\end{description}
In different works, it was shown that the performance degradation caused by the feature-type transformations can largely be mitigated. However, proposed solutions usually require a separate training step \cite{Drozdowski-DeepFaceBinarisation-ICIP-2018,Rathgeb-DeepFaceFuzzyVault-2022}. Works focusing on the performance degradation caused by the \ac{BCS} itself, have largely focused on maximizing the error correcting capabilities of the applied schemes \cite{Rathgeb-ReliabilityBalancedFusion-IJCB-2011,Tams-UnlinkableMinutiaeFuzzyVault-IET-2016}.
In this work, we examine the performance gap in the fuzzy vault scheme, a \ac{BCS} that is designed to protect integer-valued feature sets. To this end, we investigate a fuzzy vault system that employs a basic feature-type transformation approach, that does not require any training.
Our analysis identifies the variability of feature set sizes as the primary cause of performance loss in the fuzzy vault scheme, as it leads to an inconsistent error correction capability. To address this issue, we introduce an improved feature-type transformation method that achieves two key objectives: (1) it stabilizes error correction capabilities by enforcing fixed-size feature sets; (2) it allows to retain discriminative feature representations, without requiring additional training. Hence, this approach resolves the core limitations of common transformations while maintaining full compatibility with the fuzzy vault scheme. A comprehensive experimental evaluation demonstrates how the proposed approach reduces the performance loss in the fuzzy vault scheme. Our method specifically addresses two issues: (1) the performance degradation inherent to \acp{BCS}, and (2) the additional performance loss caused by feature-type transformation prior to the protection process. 
The remainder of this work is structured as follows: \autoref{sec:related_work} reviews relevant literature. Common feature-type transformations and the proposed improvements are described in \autoref{sec:method}. In \autoref{sec:results}, experiments are presented . Finally, \autoref{sec:conclusion} provides concluding remarks.
\section{Related Work}\label{sec:related_work}
\subsection{The Fuzzy Vault Scheme}\label{sec:fvs}
Prominent \acp{BCS}, like the fuzzy commitment \cite{JuelsWattenberg-FuzzyCommitmentScheme-ACM-1999} or fuzzy vault scheme \cite{JuelsSudan-FuzzyVault-IEEE-2002}, employ a key-binding approach to secure biometric templates by binding them to a secret key. This process ensures that neither the key nor the biometric feature vector can be derived from the protected template without knowledge of one of the two components, thereby satisfying the requirement of irreversibility. To account for biometric variability, error-correcting codes are integrated during the binding process. The error correcting code needs to balance the system's performance by correcting enough errors to handle intra-class variation \cite{Sutcu-FeatureTransformation-CVPRW-2008}. During verification, the system attempts to retrieve the secret key using a biometric probe. If the probe is sufficiently similar to the original template, the key can be reconstructed by correcting any remaining errors, resulting in a successful match.

Specifically, the fuzzy vault scheme introduced by \cite{JuelsSudan-FuzzyVault-IEEE-2002} encodes a secret key as coefficients of a polynomial over a finite field, then binds this polynomial to the biometric template by projecting it onto a set of integer-valued feature points during enrolment. At the time of verification, a probe template extracts a point set from the vault, which is used to reconstruct the polynomial. Ideally, all extracted points lie on the original polynomial, enabling straightforward reconstruction.
However, intra-class variations commonly cause mismatched integers—present in either the reference or probe set but not both—resulting in points that do not belong to the secret polynomial. If the overlap between the reference feature set and the probe feature set is sufficient, the secret polynomial and therefore the secret key can be recovered using polynomial reconstruction methods. The correctness of the recovered secret polynomial can be verified by computing its hash and comparing it with the stored hash-value.
Security analyses revealed that the original fuzzy vault scheme \cite{JuelsSudan-FuzzyVault-IEEE-2002} violates unlinkability and irreversibility requirements due to linkage attacks \cite{ScheirerBoult-CrackingFuzzyVaultBioEncryption-BSYM-2007,Kholmatov-CorrelationAttackFuzzyVault-SPIE-2008}. \citet{Dodi-FuzzyExtractor-ec2004} proposed an improved version of the scheme that prevents the aforementioned linkage attack, however it was shown that it is susceptible to another linkage attack based on the extended Euclidean algorithm. This vulnerability can be mitigated by applying a public random permutation to feature sets before storage and during retrieval \cite{Tams-SecurityConsiderationsFuzzyVaults-2015}. A formal description of the employed unlinkable improved fuzzy vault scheme is provided in \autoref{sec:method}.

\subsection{Feature-Type Transformations for  Biometric Cryptosystems}
\label{subsec:feature_type_transformations}
Feature-type transformations  convert biometric templates so that they are compatible with the used \ac{BCS}, in our case the fuzzy vault scheme. 
State-of-the-art biometric feature extraction methods a based on deep neural networks that extract fixed-length, real-valued feature vectors. Hence, feature-type transformations play a crucial role in converting the fixed-length, real-valued feature vectors generated by deep neural networks into data types compatible with the corresponding \ac{BCS} scheme \cite{Lim-BiometricFeatureTypeTransformation-2015}. Here, two consecutive types of feature-type transformations can be distinguished: (1) fixed-length real-valued to fixed length binary and (2) fixed-length binary to unordered integer sets.

The first category generally follows a two-step process, consisting of the feature quantisation and the feature binarisation \cite{Lim-BiometricFeatureTypeTransformation-2015}.
The feature quantisation step first maps each feature to an integer by dividing the feature space into intervals. The number of intervals determines the set of integers the features will be mapped to. A distinction is made between equal size intervals and equal probable intervals. As for the latter, each possible integer has the same probability of occurring, which maximizes the entropy of the quantified vector.
Subsequently, the binarisation step converts these integers to binary strings 
Here, two different binarisation approaches exist.  Full encoding employs all possible bit patterns for optimal space efficiency, while restricted encoding uses only a carefully selected subset of bit patterns. The latter requires more bits to represent the same integers, but may better preserve original distance relationships.
The most basic full encoding method is \ac{DBR}, which, as the name suggests, maps each integer directly to its binary representation. A more sophisticated method is the binary reflected Gray code \cite{Chang-BiometricsKeyGeneration-ICME-2004}, ensuring consecutive integers differ by exactly one bit. For restricted encoding, \ac{LSSC} best preserves the original integer-based distances after binarisation \cite{Lim-BiometricFeatureTypeTransformation-2015}.
\citet{Drozdowski-DeepFaceBinarisation-ICIP-2018} benchmarked different fixed-length real-valued to fixed-length binary  transformations for deep face recognition, showing that methods like \ac{LSSC} largely preserve the accuracy of the original templates while drastically reducing storage and computational costs. 
Transforming fixed-length real-valued vectors into unordered integer sets is less straightforward, as the extracted integer values must remain unique. While the initial steps follow the same pipeline as real-valued to binary vector transformation, an additional feature set mapping step is required to derive a set of integers from the binary vector.
\citet{Nagar-MultibiometricCryptosystemsFeauterFusion-TIFS-2012} proposed a method that segments the binary vector and converts each bit string into its decimal representation. However, this approach does not guarantee uniqueness, since different segments may map to the same integer, reducing the feature set size. Additionally, different segments can coincidentally yield the same integer during comparison.
\citet{Rathgeb-DeepFaceFuzzyVault-2022} proposed a simple method, where a binary vector is mapped to an integer set composed of each index of the binary vector that holds a 1. This method was employed in combination with different feature quantisation and binarisation methods, showing that the method works best when using 4 equal probable intervals for quantisation and \ac{LSSC} for feature binarisation. Notably, both quantisation methods, equal probable intervals and equal size intervals, require a training step to define the intervals. However, when only two intervals are used, it can be assumed that splitting the feature space at zero results in two equal probable intervals. Thus, the case of two equal probable intervals is an exception and does not require a training step.

\section{Deep Fuzzy Vault}\label{sec:method}
\begin{figure*}[h]
\centering
\includegraphics[width=0.9\linewidth]{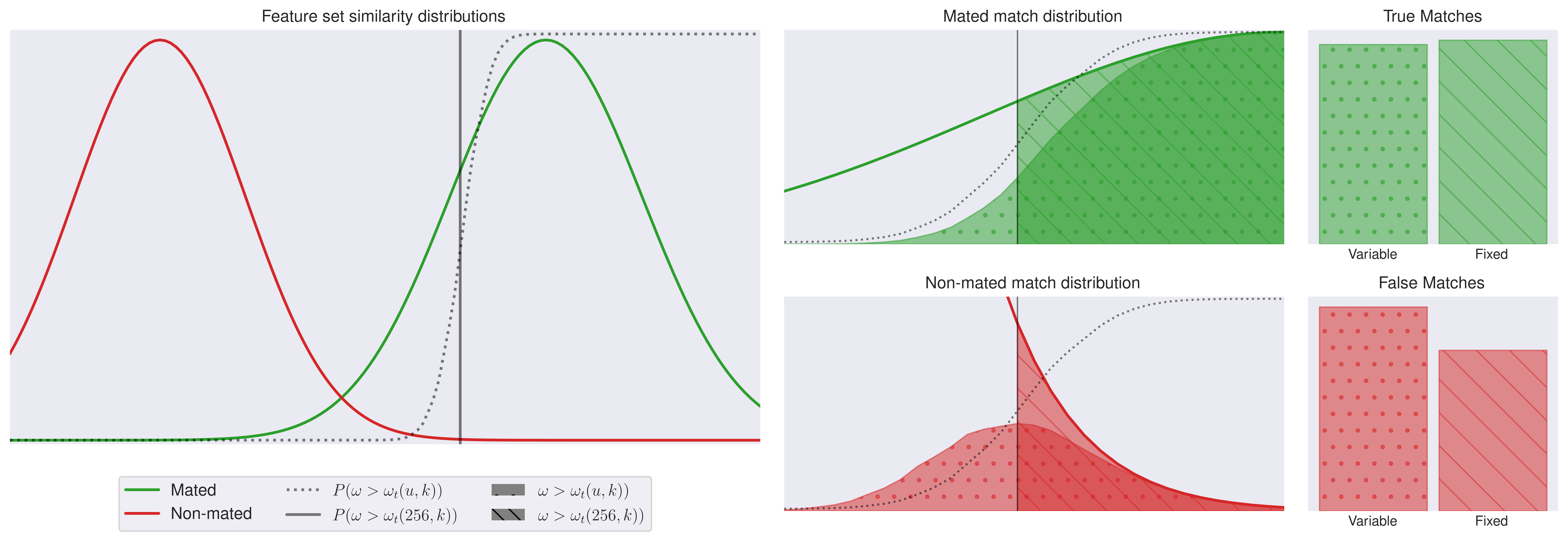}
\caption{Impact of variable versus stable thresholds on false and true match counts.
(Left) Feature set distributions with match probabilities for both threshold types.
(Center) True matches (green) and false matches (red), where dotted hatches indicate variable thresholds and dashed hatches denote stable thresholds.
(Right) Comparative bar plots of aggregate matches under both threshold conditions.}
\label{fig:FFT_explanation}
\vskip -1em
\end{figure*}
\subsection{Feature extraction}
We consider a typical fuzzy vault implementation consisting of three processing steps: feature extraction (including feature type transformation), enrolment, and verification.
The feature extraction module extracts unordered feature sets from a biometric sample. For this purpose, state-of-the-art deep learning-based feature extraction algorithms are used. These extractors produce real-valued vectors because they are trained using differentiable loss functions. Moreover, the feature vectors have a fixed length determined by the dimensionality of the corresponding deep neural network's output layer. For hardware optimization and computational efficiency, the feature vector size is often chosen as \(n=2^x\).
Since the fuzzy vault scheme is designed to protect unordered feature sets, the system includes a transformation that converts fixed-length, real-valued vectors into unordered sets of integers. As outlined in \autoref{sec:related_work}, this feature-type transformation comprises the following steps: first, each feature is quantized and binarized into $m$ bits; and finally, an unordered integer set is constructed by collecting the indices corresponding to 1s in the binary vector. Consequently, the new feature space corresponds to the power set of all possible indices.
\[
T: \mathbb{R}^{2^x} \to \mathcal{P}(\mathbb{N}^{<2^x\cdot m })
\]

As mentioned earlier, the feature transformation is a critical component of the system architecture and directly influences the performance of subsequent modules. We first provide a detailed summary of the remaining modules before outlining the design of the feature-type transformation in \autoref{sec:trafo}.

\subsection{Key Binding}
For enrolment, the improved fuzzy vault scheme is employed. This enhanced scheme reinforces the interweaving between the secret polynomial and the biometric template, thereby reducing the information leakage from the vault record. The scheme is parameterized by the maximum secret polynomial length \( k \). A higher \( k \) requires a greater degree of similarity in a comparison to produce a match, whereas a lower \( k \) permits successful key recovery with less similarity. To enrol a subject, a biometric reference is recorded and the corresponding biometric template, in the form of an integer-based feature set, is extracted.

This feature set is interpreted as a subset of a finite field \( A \subset \mathbb{F}(X) \), enabling the application of finite field arithmetic. A secret polynomial \( \kappa \) of maximum degree \( k \) is then selected. The hash \( H(\kappa) \) is computed and stored for later verification of a successful key retrieval.

To protect against record multiplicity attacks, a public bijection \( \sigma \) is applied to remap the feature set \( \hat{A} = \sigma(A) \). This bijection is deterministically derived from a public record-specific identifier, such as a user ID or hash seed. This step acts similarly to cryptographic salting, introducing pseudo-randomness while preserving performance. It ensures that vaults created from the same biometric data are non-correlatable, thereby achieving unlinkability. As shown in \cite{Tams-SecurityConsiderationsFuzzyVaults-2015,merkle-securityimprovedfuzzyvault-2014}, this mechanism effectively disrupts the overlap required for correlation or partial recovery attacks, reducing linkage probabilities to negligible levels.
For a more detailed description of unlinkability, we refer interested readers to \cite{Tams-SecurityConsiderationsFuzzyVaults-2015}, which provides a formal security analysis under this model.

The fuzzy vault \( V(X) \) is constructed by encoding the mapped set \( \hat{A} \) into a monic polynomial \( p(X) = \prod_{x \in \hat{A}} (X - x) \), followed by computing \( V(X) = \kappa(X) - p(X) \). The final record consists of the tuple \( (V(X), H(\kappa)) \), linked to the subject’s identity.

\subsection{Key Retrieval}
To verify the identity of a subject, the corresponding vault record is retrieved from the system database, yielding $V(X)$ and $H(\kappa)$. A biometric probe is then collected from the subject. The probe feature set $B$ is generated from the biometric probe using the aforementioned feature extraction pipeline, and the public bijection is applied to re-map the probe feature set to $\hat{B} = \sigma(B)$.
Using the mapped feature set, a set of probe points $P=\{(x,V(x))|x\in\hat B\}$ is computed. Since $V$ was deliberately constructed in a way that $V(x)=\kappa(x)$ for every $x \in \hat A$ every point $(x,V(x))$ with $x \in \hat B \cap \hat A$ is a genuine point, meaning that it is part of the graph of $\kappa$. 
The set of probe points $P$ is passed to the decoder. The decoder uses polynomial reconstruction to generate a list of candidate polynomials.
For each candidate polynomial $\kappa^{\prime}$, the hash $H(\kappa^{\prime})$ is computed and compared with the hash $H(\kappa)$. 
If for any candidate polynomial $H(\kappa') = H(\kappa)$, the original secret is found and the vault is unlocked. This occurs when the number of genuine points, quantified by the overlap $\omega = |\hat{B} \cap \hat{A}|$, exceeds a threshold. Theoretically, Lagrange interpolation can recover the polynomial if $\omega \geq k$. However, when the probe set size $u = |B|$ exceeds $\omega$, the decoder must iteratively test subsets of the probe set until a subset containing only genuine points is identified. This iterative process becomes computationally expensive for large set sizes, as the number of required subsets grows combinatorially. In practice, decoders often impose an iteration limit to manage this cost, which introduces a non-zero probability of failure, even when $\omega$ is theoretically sufficient.
The Guruswami-Sudan (GS) decoder \cite{Tams-UnlinkableMinutiaeFuzzyVault-IET-2016} offers a more efficient polynomial reconstruction.
The approach is able to recover the secret $\kappa$ under the condition that $\omega\ge\sqrt{u\cdot(k-1)}$ in polynomial time. 
This is achieved by interpolating a specific bivariant polynomial $Q(X,Y)$ that has at most a $(1,k-1)$ weighted degree of $l$ and satisfies $Q(X+x,Y+y)$ having a multiplicity of at least $\mu$ for every point $(x,y)\in P$.
By factoring $Q(X,Y)$ into univariant polynomials $\kappa^{\prime}$ with polynomial degrees of at most $k$, a list of candidate polynomials is created. For a more detailed explanation, refer to \cite{Guruswami-ImprovedDecodingOfReedSolomon-1998}.
It has been shown that for sufficiently large values of $l$ and $\mu$ the correct polynomial can be recovered if the point set contains at least $\omega>\sqrt{u\cdot(k-1)}$ genuine points.
Note that since \(\sigma\) is a bijection, \(| B \cap A| = |\hat B \cap \hat A|\) and \(|B| = |\hat B|\), meaning that both \(u\) and \(\omega\) are unaffected by the permutation. Therefore, the application of the permutation \(\sigma\) does not alter the performance of the scheme.

\subsection{Feature-Type Transformation}\label{sec:trafo}
As previously mentioned, the feature-type transformation maps a fixed-length, real-valued vector \(v=(x_0,\dots,x_{n})\) to a variable-size integer-valued feature set \(A\subset\mathbb{N}^{<n\cdot m}\) through three consecutive steps: \emph{feature quantization}, \emph{binarization}, and \emph{set mapping}.
Since methods for feature quantization that use more than two intervals typically require additional training, we consider a the following approach as a baseline: dividing the feature space into two equal-probable intervals. 
Because feature distributions are balanced around zero due to the training of the feature extractor, this can be achieved by simply dividing the feature space at zero.
Therefore, each feature that is zero or greater is mapped to a 1, and every feature less than zero is mapped to a 0:
\begin{gather*}
    T_Q((x_0,\dots,x_{n}))=(q(x_0),\dots,q(x_{n}))\\
    q(x_i) = \begin{cases} 
        1 & \text{if } x_i \geq 0 \\ 
        0 & \text{otherwise.}
    \end{cases}.
\end{gather*}
As using only two intervals for quantization already yields a binary vector, the feature binarization \( T_B \) using \ac{DBR} is trivial, as it just maps each decimal 0 to a binary 0 and each decimal 1 to a binary 1.
The feature set mapping operation \( T_M \) subsequently constructs the feature set by collecting indices of all 1-valued features:
\begin{gather*}
    T_M((x_0^B,\dots,x_{n-1}^B)) = \{i \mid 0 \leq i < n, x_i^B = 1\}.
\end{gather*}
As an example, consider the feature vector \( v \) of length \( n = 8 \). 
Using equal-probable quantization, each feature is mapped to 1 if it is non-negative and to 0 otherwise:
\begin{align*}
v &= (-0.6, 0.3, 1.2, -0.9, 0.5, -1.1, 0.7, 0.1) \\
T_Q(v) = v_B &= (0,\ 1,\ 1,\ 0,\ 1,\ 0,\ 1,\ 1)
\end{align*}
The binary vector \( v_B \) is then mapped to a feature set by collecting the indices of all 1-valued entries:
\begin{align*}
T_M(v_B) = A &= \{1,\ 2,\ 4,\ 6,\ 7\}
\end{align*}

For a feature vector of length \( n = 512 \), the transformation yields feature sets with an average size of 256 and a standard deviation of approximately \(11.31\). The matching decision depends on both the overlap \(\omega\) and the probe set size \(u\). To properly evaluate the impact of threshold variability, we analyze the relative similarity \(\omega/u\). The GS decoder succeeds if \(\omega > \sqrt{u(k-1)}\), resulting in a size-dependent threshold \(\omega_t(u,k) = \lceil \sqrt{u(k-1)} \rceil\). While \(\omega\) scales linearly with \(u\), \(\omega_t\) grows sublinearly due to its dependence on \(\sqrt{u}\), leading to lower relative thresholds \(\omega_t(u,k)/u\) for larger sets. This increases the probability of false accepts and negatively affects system performance, as illustrated in \autoref{fig:FFT_explanation}. The figure's left panel shows the feature set similarity distribution \(\omega/u\) for mated and non-mated comparisons, along with the threshold for the average set size \(\omega_t(256,k)/256\) and the match probability \(P(\omega \geq \omega_t(u,k))\). The detailed plots compare fixed thresholds against variable thresholds, which are cumnmulated in the respective bar plots, revealing that fixed thresholds marginally improve true matches while significantly reducing false matches.
As a concrete example, consider a system using polynomial degree \(k=96\) and a non-mated comparison with probe set size \(u=264\) and observed overlap \(\omega=159\). The required threshold is calculated as \(\omega_t(264,96) = \lceil \sqrt{264 \times 95} \rceil = 159\), thus resulting in a false match, since \(\omega \geq \omega_t\). Now consider the same comparison assuming a fixed probe set size of \(u=256\), where the overlap scales proportionally to \(\omega' = \omega \cdot \frac{256}{264} \approx 154\). The new threshold is \(\omega_t(256,96) = \lceil \sqrt{256 \times 95} \rceil = 156\). Since \(\omega' = 154 < 156 = \omega_t\), the comparison correctly results in a rejection. This example demonstrates how fixing the set size stabilizes the decision boundary and helps prevent false accepts in borderline cases.
While this analysis makes the simplifying assumption that fixing the feature set size does not substantially alter the similarity distribution, in practice, any changes are minimal because the variation in set size is relatively small. Thus, the core conclusion remains: variable thresholds degrade system performance by simultaneously reducing true matches and increasing false accepts.
This motivates an important additional requirement for fuzzy vault schemes: the feature transformation shall produce sets of fixed size. Formally, the transformation should map all input vectors to integer sets of identical cardinality. This necessitates that the binarized vectors, prior to set mapping, exhibit a predetermined Hamming weight. While this could theoretically be achieved by either enforcing uniform usage of quantized integers or designing binarization steps with fixed Hamming weight outputs, we prioritize the former approach for its flexibility and compatibility with deep feature characteristics.
Deep learning-based feature vectors are particularly suitable for this strategy due to two inherent properties. First, normalization layers used during training help ensure that the extracted feature vectors have values that are roughly centered around zero with consistent variation across all dimensions. Second, their high dimensionality (\(n \gg 1\)) guarantees sufficient intra-vector value diversity, even within individual samples.
These properties enable equal-frequent quantization, a method that adapts interval boundaries per feature vector, analogous to equal-probable intervals but without requiring population-level statistics. Concretely, our quantization approach operates in three stages: (1) ranking the features of the input vector, (2) dividng the ranked features into \(m\) equal-sized quantiles (each containing \(n/m\) features), and (3) assigning integer labels sequentially. Here, \(\text{rank}(x_i)\) denotes the index of the feature \(x_i\) when sorting the full vector in ascending order, assigning rank 0 to the smallest and \(n-1\) to the largest element.
This procedure ensures that exactly \(n/m\) features map to each integer, thereby fixing the Hamming weight of intermediate binary representations. The resulting feature sets inherit this size consistency, eliminating variability in the fuzzy vault's probe set size \(u\) and stabilizing the minimum required overlap \(\omega_t\):
\begin{gather*}
    T_Q((x_0,\dots,x_{n}))=(q(x_0),\dots,q(x_{n}))\\
    q(x_i) = \left \lfloor\frac{\text{rank}(x_i)\cdot  m}{n} \right \rfloor
\end{gather*}

Choosing an interval count of \( m = 2 \), we can apply the same binarization and set mapping steps as in the aforementioned baseline transformation.
As an example we will consider the same vector $v$ of length $n=8$ as before. 
First, we rank the feature values in ascending order. Each rank corresponds to the position of the value in the sorted vector, starting from 0 for the smallest:
\begin{align*}
\text{ranks}(v) &= (2,\ 4,\ 7,\ 1,\ 5,\ 0,\ 6,\ 3)
\end{align*}
The lowest 4 ranks (0–3) are mapped to interval 0, and the highest 4 (4–7) to interval 1:
\begin{align*}
T_Q(v) &= (0,\ 1,\ 1,\ 0,\ 1,\ 0,\ 1,\ 0)
\end{align*}
Again, the feature set is formed by collecting indices of all 1-valued entries:
\begin{align*}
T_M(T_Q(v)) = A &= \{1,\ 2,\ 4,\ 6\}
\end{align*}  
Notably, this set differs from the one in the first example only in that index 7 (corresponding to the feature closest to zero) is no longer included.
This method allows for training-free quantization with more than two intervals while maintaining consistent set sizes. In this work, we evaluate two configurations using the equal-frequent quantization approach: one using \(m = 2\) intervals and \ac{DBR}, to be compared with a baseline using the equal-probable method also using \ac{DBR}, and one mimicking the best-performing configuration in \cite{Rathgeb-DeepFaceFuzzyVault-2022}, which uses \(m = 4\) quantization levels and binarization via \ac{LSSC}.

\section{Empirical evaluation}\label{sec:results}
\subsection{Experimental Setup}
\label{sec:experiments}
\begin{figure}
    \centering
    \begin{subfigure}[b]{0.4\columnwidth}
        \centering
    \includegraphics[width=.48\linewidth]{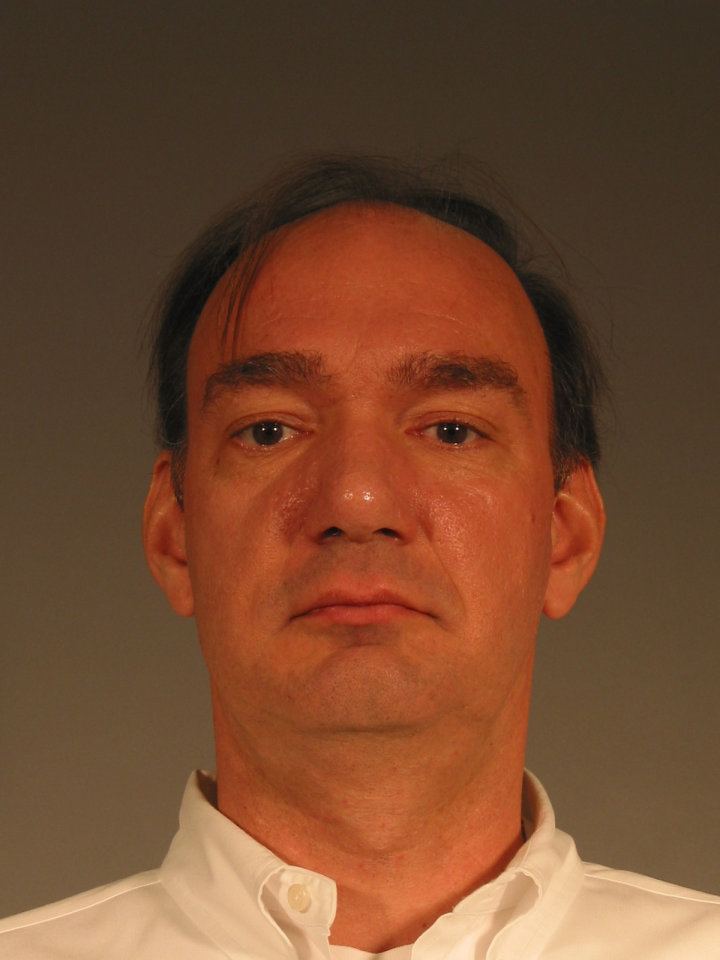}
    \includegraphics[width=.48\linewidth]{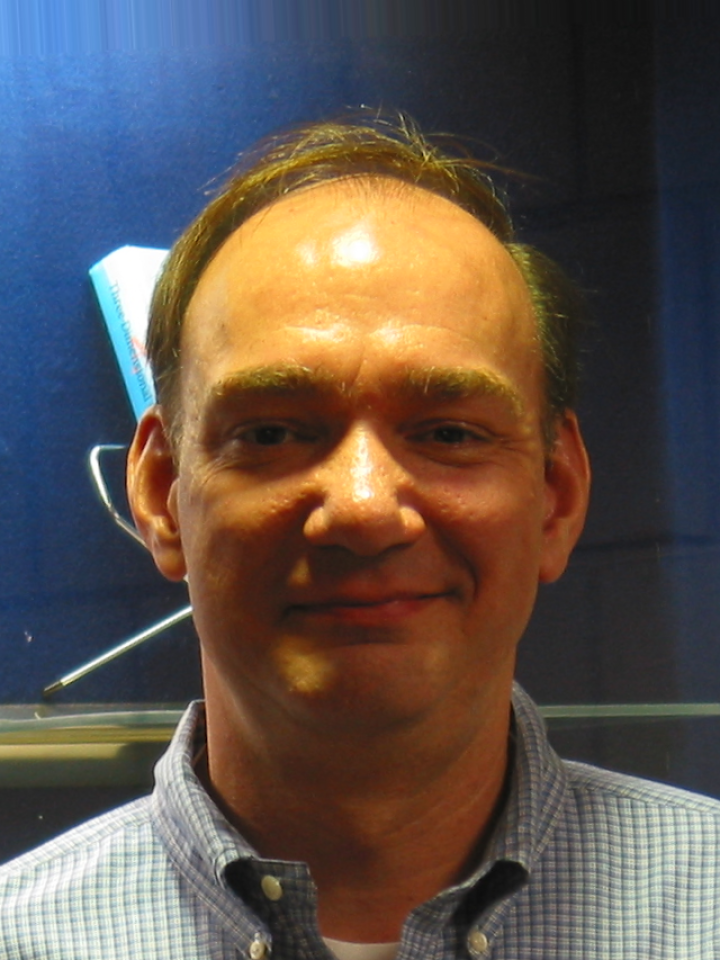}
    \caption{FRGC}
    \end{subfigure}\hfill
    \begin{subfigure}[b]{0.4\columnwidth}
    \centering
    \includegraphics[width=.4\linewidth]{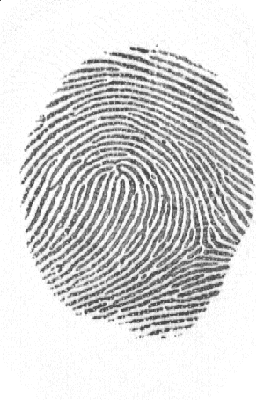}
    \includegraphics[width=.4\linewidth]{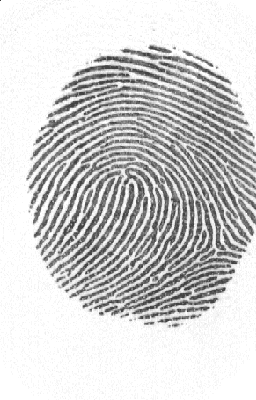}
    \caption{MCYT}
    \end{subfigure}\vspace{0.2cm}
    \begin{subfigure}[b]{0.45\columnwidth}
    \centering 
    \includegraphics[width=0.45\linewidth]{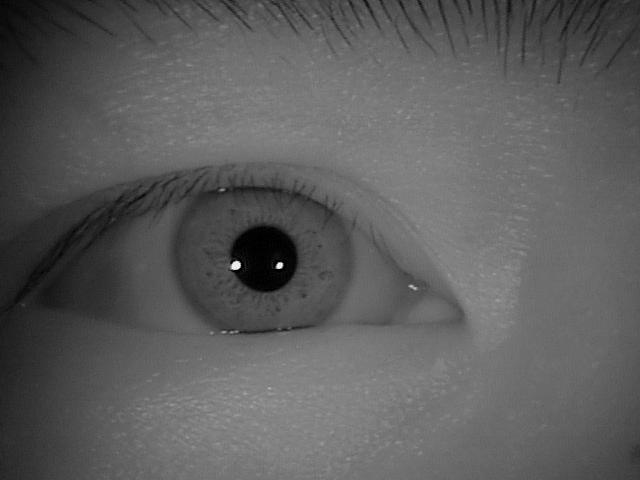}
\includegraphics[width=0.45\linewidth]{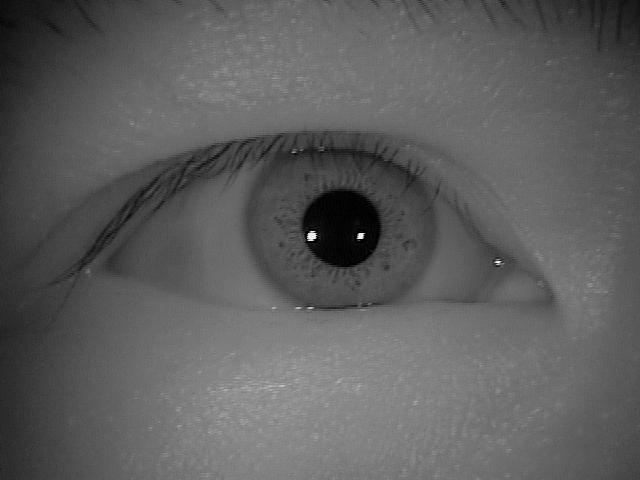}
    \caption{CASIA}
    \end{subfigure}
    \caption{Example images for the used datasets.}
    \label{fig:frgc}
    \vskip -1em
\end{figure}
A series of experiments will be conducted using the system described in \autoref{sec:method}, using the three mentioned feature type transformation setups: (1) 2 equal-probable intervals and \ac{DBR}; (2) 2 equal-frequent intervals and \ac{DBR}; (3) 4 equal-frequent intervals and \ac{LSSC}. We analyze performance differences in between the feature-type transformation approaches using 2 quantisation intervals to assess whether fixing the feature set size eliminates these variations. Using the third transformation approach, we evaluate the applicability of our proposed quantisation method with multiple intervals ($>$2) and measure its performance impact.
The experiments will be performed on face, fingerprint, and iris features to evaluate the effectiveness of the improved transformation. For each characteristic, both the unprotected and transformed systems will also be analysed to assess the performance gap. 
The systems will be empirically tested against mated and non-mated comparisons, computing \ac{GMR} and \ac{FMR} to assess the biometric system performance. Additionally, the \ac{FAS} is calculated as a security metric. The \ac{FAS} quantifies resistance against false accept attacks, which represent the most significant threat to \ac{BCS} when other attack vectors are unavailable. In such an attack, an adversary systematically submits non-mated samples to induce a false match, potentially compromising the system by revealing the cryptographic key and gaining access to protected biometric data. This attack model assumes the adversary has full knowledge of the algorithm, access to protected templates, and statistical data about the biometric feature vectors.
The \ac{FAS} represents the expected number of non-mated comparisons needed to produce a false match multiplied by the computational cost $t$ of one comparison, where $t$ is the ratio of the runtime of a polynomial reconstruction using the Guruswami-Sudan decoder to the runtime of a single Lagrange iteration. The \ac{FAS} in bits is computed as $\text{FAS} = \log_2(t \cdot \log(0.5)/\log(1 - \text{FMR}))$ \cite{Rathgeb-BTP-Survey-EURASIP-2011}.
For face recognition, we used a subset of FRGCv2, which includes frontal images captured under both controlled and uncontrolled conditions. Feature extraction was performed using a pretrained model from \cite{kim-adafacequalityadaptivemargin-2023}, optimized for recognition accuracy through quality-adaptive margins. 
Iris samples were sourced from the CASIA-Iris-Thousand dataset. To ensure consistency, left and right irises were treated as separate identities. Features were extracted using the approach from \cite{Boutros-LowResolutionIris-BIOSIG-2022}, employing a pretrained model as described in \cite{OsorioRoig-PrivacyPreservingMultiBiometricIndexing-TIFS-2024}.  
For fingerprint recognition, we used samples collected from multiple fingers per subject. As there is evidence that fingerprints from different fingers of the same individual exhibit higher similarity than those from different subjects \cite{Guo-UnveilingIntraPersonFingerprintSimilarity-2024}, we excluded such samples from non-mated comparisons. Feature extraction was performed using DeepPrint \cite{Engelsma-LearningFixedLengthFingerprintRepresentation-2019}, specifically the texture branch implementation from \cite{Rohwedder-FixedLengthFingerprintDNN-BIOSIG-2023}.
\subsection{Results}
\begin{figure*}[t]
    \centering
    \begin{subfigure}[b]{0.3\textwidth}
        \centering
        \includegraphics[width=\linewidth]{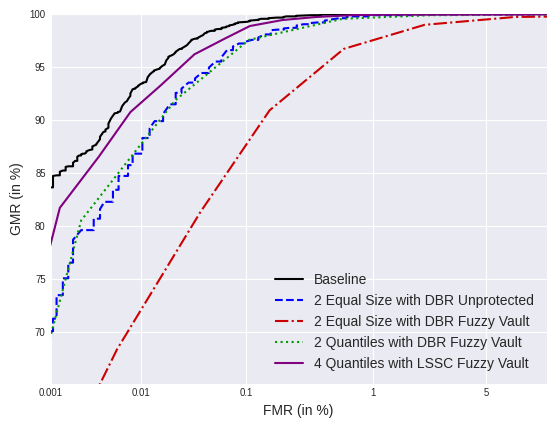}
        \caption{Face recognition}
        \label{fig:face-performance}
    \end{subfigure}
    \begin{subfigure}[b]{0.3\textwidth}
        \centering
        \includegraphics[width=\linewidth]{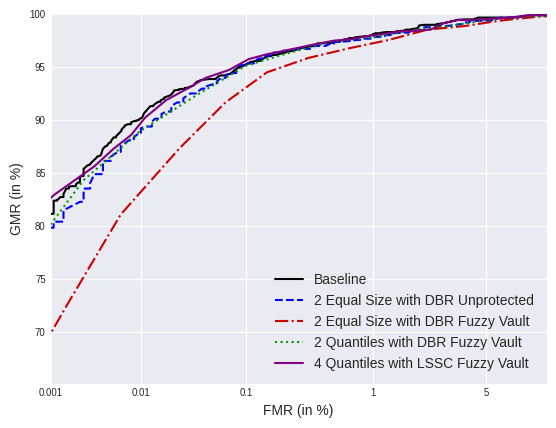}
        \caption{Fingerprint recognition}
        \label{fig:fingerprint-performance}
    \end{subfigure}
    \begin{subfigure}[b]{0.3\textwidth}
        \centering
        \includegraphics[width=\linewidth]{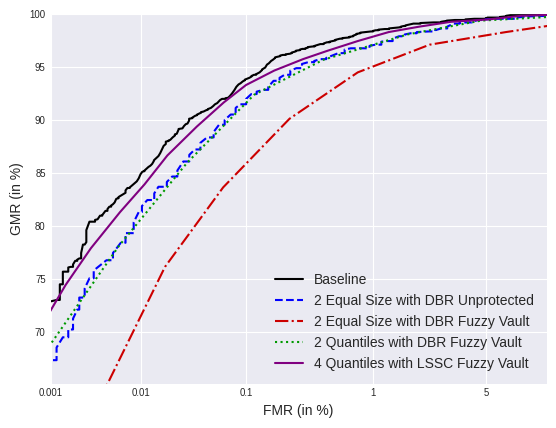}
        \caption{Iris recognition}
        \label{fig:iris-performance}
    \end{subfigure}
    \caption{ROC curves comparing system performance across biometric modalities: (a) face, (b) fingerprint, and (c) iris recognition systems showing performance of unprotected, binarised, and transformed systems.}
    \label{fig:all-performance}
    \vskip -1em 
\end{figure*}

The results will be presented in two parts: First, the results of the experiments involving only 2 intervals are discussed for each characteristic, assessing how the proposed approach compares to the naive approach and also assessing the performance gap during template protection. Second, the results of the experiments using 4 \textit{equal-frequent intervals} will be presented, assessing the performance gap caused by the complete system (involving feature type transformation and template protection). 
The results of the experimental evaluation are displayed in tables, additionally the results are illustrated in \autoref{fig:all-performance} using ROC plots to visualize performance gaps.
\begin{table}[h]
\centering
\scriptsize
\caption{Face fuzzy vault performance using 2 intervals.}
\label{tab:face}
\begin{tabular}{l ccc@{\hskip 1em}ccc} \toprule
 & \multicolumn{3}{c}{2 Equal Probable} & \multicolumn{3}{c}{2 Equal Frequent} \\ 
 \cmidrule(lr{1em}){2-4} \cmidrule(lr{1em}){5-7}
$k$ & \makecell{GMR\\ (in \%)} & \makecell{FMR\\ (in \%)} & \makecell{FAS\\ (in bit)} & \makecell{GMR\\ (in \%)} & \makecell{FMR\\ (in \%)} & \makecell{FAS\\ (in bit)} \\ \midrule
16 & 100.00 & 100.0000 & 0.00 & 100.00 & 100.0000 & 0.00 \\ 
32 & 100.00 & 99.9983 & 3.32 & 100.00 & 99.9999 & 2.81 \\ 
48 & 100.00 & 98.6518 & 3.32 & 100.00 & 99.5727 & 2.63 \\ 
64 & 100.00 & 70.1352 & 4.53 & 100.00 & 69.8698 & 4.31 \\ 
80 & 99.83 & 19.1256 & 6.61 & 99.94 & 9.8791 & 6.95 \\ 
96 & 98.98 & 2.1956 & 8.89 & 99.55 & 0.6063 & 10.70 \\ 
112 & 90.90 & 0.1603 & 12.28 & 91.87 & 0.0221 & 15.17 \\ 
128 & 68.33 & 0.0056 & 16.79 & 65.78 & 0.0007 & 19.69 \\ 
144 & 42.75 & 0.0002 & 21.21 & 37.75 & 0.0001 & 22.21 \\ 
\bottomrule
\end{tabular}

\end{table}
The face recognition results in \autoref{tab:face} demonstrate that the proposed equal-frequent quantization outperforms the naive equal-probable intervals approach. At $k=96$, the GMR improves by 0.57 percentage points (from 98.98\% to 99.55\%), and the \ac{FAS} increases by 1.89 bits (from 8.89 to 10.70 bits). \autoref{fig:face-performance} shows that the ROC curve of the fuzzy vault using the proposed transformation nearly overlaps that of the unprotected transformed system, confirming that the performance gap introduced by template protection is effectively mitigated.
\begin{table}[h]
\centering
\scriptsize
\caption{Fingerprint fuzzy vault performance using 2 intervals.}
\label{tab:finger}
\begin{tabular}{l ccc@{\hskip 1em}ccc} \toprule
 & \multicolumn{3}{c}{2 Equal Probable} & \multicolumn{3}{c}{2 Equal Frequent} \\ 
 \cmidrule(lr{1em}){2-4} \cmidrule(lr{1em}){5-7}
$k$ & \makecell{GMR\\ (in \%)} & \makecell{FMR\\ (in \%)} & \makecell{FAS\\ (in bit)} & \makecell{GMR\\ (in \%)} & \makecell{FMR\\ (in \%)} & \makecell{FAS\\ (in bit)} \\ \midrule
16 & 100.00 & 99.9991 & 5.46 & 100.00 & 99.9998 & 5.23 \\ 
32 & 100.00 & 98.3203 & 4.44 & 100.00 & 98.9332 & 4.51 \\ 
48 & 99.94 & 79.1125 & 4.82 & 100.00 & 81.1213 & 4.75 \\ 
64 & 99.94 & 47.3122 & 5.49 & 99.94 & 48.3503 & 4.91 \\ 
80 & 99.94 & 25.2776 & 6.17 & 99.94 & 24.2825 & 6.04 \\ 
96 & 99.94 & 12.7996 & 6.22 & 99.94 & 12.3347 & 6.27 \\ 
112 & 99.37 & 5.9353 & 7.01 & 99.49 & 5.3140 & 7.21 \\ 
128 & 98.52 & 2.2845 & 8.10 & 98.47 & 1.8318 & 8.39 \\ 
144 & 96.76 & 0.6776 & 9.60 & 97.38 & 0.4977 & 10.03 \\ 
160 & 94.49 & 0.1518 & 11.47 & 95.11 & 0.1001 & 12.06 \\ 
176 & 87.32 & 0.0242 & 13.81 & 88.86 & 0.0095 & 15.16 \\ 
192 & 72.83 & 0.0016 & 17.30 & 76.29 & 0.0004 & 19.21 \\ 
\bottomrule
\end{tabular}

\end{table}
For fingerprints (\autoref{tab:finger}) at $k=128$, GMR remains stable (98.47\% vs 98.52\% at $k=128$), while \ac{FAS} increases by 0.77 bits (7.19 to 7.96 bits). \autoref{fig:fingerprint-performance} reveals that the performance gap is virtually eliminated, mirroring the face recognition results. The marginal GMR difference suggests the proposed method enhances security without compromising accuracy.
\begin{table}[h]
\centering
\scriptsize
\caption{Iris fuzzy vault performance using 2 intervals.}
\label{tab:iris}
\begin{tabular}{l ccc@{\hskip 1em}ccc} \toprule
 & \multicolumn{3}{c}{2 Equal Probable} & \multicolumn{3}{c}{2 Equal Frequent} \\ 
 \cmidrule(r{1em}){2-4} \cmidrule{5-7}
$k$ & \makecell{GMR\\ (in \%)} & \makecell{FMR\\ (in \%)} & \makecell{FAS\\ (in bit)} & \makecell{GMR\\ (in \%)} & \makecell{FMR\\ (in \%)} & \makecell{FAS\\ (in bit)} \\ \midrule
16 & 100.00 & 100.0000 & 0.00 & 100.00 & 100.0000 & 0.00 \\ 
32 & 100.00 & 99.9095 & 3.60 & 100.00 & 99.9885 & 3.50 \\ 
48 & 100.00 & 92.3167 & 4.11 & 100.00 & 95.9239 & 3.43 \\ 
64 & 100.00 & 53.4109 & 5.20 & 100.00 & 57.6746 & 5.08 \\ 
80 & 99.43 & 15.1143 & 6.25 & 99.83 & 12.7706 & 6.80 \\ 
96 & 97.10 & 2.3579 & 8.70 & 98.07 & 1.7084 & 9.21 \\ 
112 & 90.11 & 0.2346 & 11.70 & 92.38 & 0.1184 & 12.71 \\ 
128 & 76.12 & 0.0174 & 15.10 & 78.06 & 0.0058 & 16.72 \\ 
144 & 54.97 & 0.0011 & 18.89 & 56.11 & 0.0001 & 22.21 \\ 
\bottomrule
\end{tabular}

\end{table}
Iris recognition (\autoref{tab:iris}) exhibits a similar trend: at $k=96$, GMR improves from 97.10\% to 98.07\% and \ac{FAS} gains 0.58 bits (7.91 to 8.49 bits). The ROC curve in \autoref{fig:iris-performance} demonstrates near-identical performance between the improved transformation and the binarised system, closing the performance gap entirely.
\begin{table}[h]
    \caption{Results using 4 equal-frequent intervals and \ac{LSSC}.}
\begin{adjustbox}{max width=\linewidth}
    \centering
    \label{tab:4eqfLSSC}
    \begin{tabular}{l ccc@{\hskip 1em}ccc@{\hskip 1em}ccc} \toprule
 & \multicolumn{3}{c}{Iris} & \multicolumn{3}{c}{Face} & \multicolumn{3}{c}{Fingerprint} \\ 
 \cmidrule(r{1em}){2-4} \cmidrule(r{1em}){5-7} \cmidrule{8-10}
$k$ & \makecell{GMR\\ (in \%)} & \makecell{FMR\\ (in \%)} & \makecell{FAS\\ (in bit)} & \makecell{GMR\\ (in \%)} & \makecell{FMR\\ (in \%)} & \makecell{FAS\\ (in bit)} & \makecell{GMR\\ (in \%)} & \makecell{FMR\\ (in \%)} & \makecell{FAS\\ (in bit)} \\ \midrule
288 & 100 & 21.9791 & 5.35 & 100 & 20.9293 & 5.35 & 100 & 31.8603 & 3.89 \\ 
320 & 99 & 5.1596 & 6.91 & 100 & 2.5360 & 7.92 & 100 & 19.3227 & 4.06 \\ 
352 & 97 & 0.7814 & 9.44 & 99 & 0.2057 & 11.39 & 100 & 11.0517 & 4.71 \\ 
384 & 93 & 0.1003 & 12.23 & 93 & 0.0159 & 14.89 & 100 & 6.0435 & 5.39 \\ 
416 & 84 & 0.0108 & 15.20 & 76 & 0.0009 & 18.88 & 99 & 2.8882 & 6.18 \\ 
448 & 70 & 0.0007 & 18.88 & 53 & 0.0002 & 20.68 & 98 & 1.1943 & 7.26 \\ 
480 & 52 & 0.0000 & 0.00 & 32 & 0.0000 & 0.00 & 97 & 0.3984 & 9.04 \\ 
512 & 38 & 0.0000 & 0.00 & 21 & 0.0000 & 0.00 & 96 & 0.1059 & 10.67 \\ 
544 & 24 & 0.0000 & 0.00 & 13 & 0.0000 & 0.00 & 92 & 0.0179 & 13.01 \\ 
576 & 12 & 0.0000 & 0.00 & 5 & 0.0000 & 0.00 & 86 & 0.0032 & 15.31 \\ 
\bottomrule
\end{tabular}

\end{adjustbox}
\end{table}
Finally, the results for the transformation using \textit{4 equal-frequent intervals}, which theoretically retains more discriminative information and should therefore improve performance, are displayed in \autoref{tab:4eqfLSSC}, grouping all characteristics together. For face recognition at \( k=352 \), this approach achieved 99\% \ac{GMR} with 11.09 bits \ac{FAS}, while fingerprint recognition at \( k=496 \) maintained 97\% \ac{GMR} with 10.04 bits \ac{FAS}, and iris recognition at \( k=336 \) reached 99\% \ac{GMR} with 8.30 bits \ac{FAS}. The \ac{ROC} plots in \autoref{fig:all-performance} demonstrate that this method further reduces the performance gap, surpassing the naive binarisation approach and showing only minimal residual gaps for face and iris while completely closing the gap for fingerprints.  
However, when comparing these \ac{FAS} values between the 2-interval and 4-interval approaches, the improvements appear modest at first glance: face recognition shows 11.09 bits versus 10.70 bits, fingerprints improve from 7.96 to 10.04 bits, while iris recognition decreases slightly from 8.49 to 8.30 bits. These comparisons require careful interpretation because the \ac{FAS} calculation uses the relative runtime \( t \), which normalizes the Guruswami-Sudan decoder's runtime by the duration of a single Lagrange interpolation at the given polynomial degree. This normalization means that while the \ac{FAS} values may appear similar, the actual computational effort required for attacks differs substantially due to how the Lagrange runtime scales with polynomial degree.
For face recognition, the 4-interval method at \( k=352 \) has a Lagrange runtime of 4,755 µs compared to 331 µs for the 2-interval method at \( k=96 \), a 14 times increase. Although the \ac{FAS} difference is only 0.39 bits, the higher Lagrange runtime means each bit of security requires more computational effort from an attacker. The same pattern holds for fingerprints, where the Lagrange runtime increases from 587 µs at \( k=128 \) to 8,134 µs at \( k=496 \) , and for iris recognition, where it grows from 331 µs at \( k=96 \) to 4,522 µs at \( k=336 \).
\subsection{Discussion}
Our quantisation method assumes moderate consistency across feature dimensions, typically supported by normalisation layers in deep networks. Since the method operates on feature rankings rather than absolute values, it remains effective without strict centring or uniform variance. We observe stable performance even under moderate deviations. This behaviour supports the method’s generality, though further study on how network architectures and data shifts affect this assumption remains an important direction for future work.
The experimental results demonstrate that enforcing fixed-size feature sets in \ac{BCS} effectively eliminates the performance gap caused by variable feature set sizes across the three biometric modalities: face, fingerprint, and iris. The proposed \textit{equal-frequent quantisation} method consistently outperformed the naive approach by stabilising the error correction capabilities of the decoder, confirming that performance loss in fuzzy vault schemes primarily stems from unstable similarity thresholds due to varying set sizes. 
The \ac{ROC} curves in Figure~\ref{fig:all-performance} show that template protection introduces negligible performance loss when using our fixed-size transformation. Although the 4-interval transformation achieved similar \ac{FAS} values to the 2-interval approach, it demonstrated improved trade-offs between \ac{GMR} and \ac{FMR}, highlighting limitations in \ac{FAS} comparability across different polynomial degrees. While this work focuses on bridging the performance gap between protected and unprotected systems, the computational cost of decoding increases with polynomial degree and thus with higher security. Although this increase is acceptable in our setting, real-time deployments with strict \ac{FMR} requirements may require further optimization, which we leave for future work.
More importantly, we demonstrated that \textit{equal-frequent interval} quantisation can nearly close the performance gap completely while enabling training-free quantisation into arbitrary numbers of intervals.
These findings prove that careful feature space quantisation can maintain template security while achieving recognition performance comparable to unprotected systems. The consistent results across three fundamentally different biometric characteristics suggest broad applicability of our approach to the fuzzy vault scheme.

\section{Conclusion}\label{sec:conclusion}
This work demonstrates that performance degradation in fuzzy vault schemes stems primarily from variable feature set sizes, which introduce instability in error correction thresholds. By proposing an equal-frequent quantisation method that enforces fixed-size sets, we bridge the gap between cryptographic security and recognition accuracy. Experiments across face, fingerprint, and iris modalities confirm that our approach achieves: (1) near-identical performance as unprotected systems, (2) training-free compatibility with arbitrary quantisation intervals, (3) minimizing the performance gap associated with \ac{BCS}. Crucially, the method’s consistency across diverse biometric traits underscores its generalizability. Future work will explore adaptive set-size optimization and decoder acceleration to further reduce computational overhead. These advances solidify fuzzy vaults as a practical solution for biometric template protection, balancing security and usability without sacrificing performance.
\newpage
\bibliographystyle{IEEEtranN}
\bibliography{IEEEabrv,biometrics,local_biometrics}

\begin{thebibliography}{26}
\providecommand{\natexlab}[1]{#1}
\providecommand{\url}[1]{#1}
\csname url@samestyle\endcsname
\providecommand{\newblock}{\relax}
\providecommand{\bibinfo}[2]{#2}
\providecommand{\BIBentrySTDinterwordspacing}{\spaceskip=0pt\relax}
\providecommand{\BIBentryALTinterwordstretchfactor}{4}
\providecommand{\BIBentryALTinterwordspacing}{\spaceskip=\fontdimen2\font plus
\BIBentryALTinterwordstretchfactor\fontdimen3\font minus \fontdimen4\font\relax}
\providecommand{\BIBforeignlanguage}[2]{{%
\expandafter\ifx\csname l@#1\endcsname\relax
\typeout{** WARNING: IEEEtranN.bst: No hyphenation pattern has been}%
\typeout{** loaded for the language `#1'. Using the pattern for}%
\typeout{** the default language instead.}%
\else
\language=\csname l@#1\endcsname
\fi
#2}}
\providecommand{\BIBdecl}{\relax}
\BIBdecl

\bibitem[{ISO/IEC JTC1 SC27 Security Techniques}(2022)]{ISO-IEC-24745-TemplateProtection-2022}
{ISO/IEC JTC1 SC27 Security Techniques}, \emph{{ISO/IEC} 24745:2022. Information Technology - Security Techniques - Biometric Information Protection}, International Organization for Standardization, 2022.

\bibitem[Uludag et~al.(2004)Uludag, Pankanti, Prabhakar, and Jain]{Uludag-BiometricCryptosystems-IEEE-2004}
U.~Uludag, S.~Pankanti, S.~Prabhakar, and A.~Jain, ``Biometric cryptosystems: Issues and challenges,'' \emph{Proc. of the {IEEE}}, vol.~92, no.~6, pp. 948--960, May 2004.

\bibitem[Nandakumar and Jain(2015)]{NandakumarJain-BiometricTemplateProtection-SPM-2015}
K.~Nandakumar and A.~Jain, ``Biometric template protection: Bridging the performance gap between theory and practice,'' \emph{Signal Processing Magazine, {IEEE}}, vol.~32, no.~5, pp. 88--100, September 2015.

\bibitem[Drozdowski et~al.(2018)Drozdowski, Struck, Rathgeb, and Busch]{Drozdowski-DeepFaceBinarisation-ICIP-2018}
P.~Drozdowski, F.~Struck, C.~Rathgeb, and C.~Busch, ``Benchmarking binarisation schemes for deep face templates,'' in \emph{Intl. Conf. on Image Processing ({ICIP})}.\hskip 1em plus 0.5em minus 0.4em\relax IEEE, October 2018, pp. 191--195.

\bibitem[Rathgeb et~al.(2022)Rathgeb, Merkle, Scholz, B.~Tams, and Nesterowicz]{Rathgeb-DeepFaceFuzzyVault-2022}
C.~Rathgeb, J.~Merkle, J.~Scholz, B.~B.~Tams, and V.~Nesterowicz, ``Deep face fuzzy vault: Implementation and performance,'' \emph{Computers \& Security}, vol. 113, p. 102539, Feb. 2022.

\bibitem[Rathgeb et~al.(2011)Rathgeb, Uhl, and Wild]{Rathgeb-ReliabilityBalancedFusion-IJCB-2011}
C.~Rathgeb, A.~Uhl, and P.~Wild, ``Reliability-balanced feature level fusion for fuzzy commitment scheme,'' in \emph{2011 International Joint Conference on Biometrics (IJCB)}, 2011, pp. 1--7.

\bibitem[Tams(2016)]{Tams-UnlinkableMinutiaeFuzzyVault-IET-2016}
B.~Tams, ``Unlinkable minutiae-based fuzzy vault for multiple fingerprints,'' \emph{IET Biometrics}, vol.~5, no.~3, pp. 170--180, 2016.

\bibitem[Juels and Wattenberg(1999)]{JuelsWattenberg-FuzzyCommitmentScheme-ACM-1999}
A.~Juels and M.~Wattenberg, ``A fuzzy commitment scheme,'' in \emph{6th {ACM} Conf. on Computer and Communications Security}, 1999, pp. 28--36.

\bibitem[Juels and Sudan(2002)]{JuelsSudan-FuzzyVault-IEEE-2002}
A.~Juels and M.~Sudan, ``A fuzzy vault scheme,'' in \emph{{IEEE} Intl. Symposium on Information Theory}, June 2002, p. 408.

\bibitem[Sutcu et~al.(2008)Sutcu, Rane, Yedidia, Draper, and Vetro]{Sutcu-FeatureTransformation-CVPRW-2008}
Y.~Sutcu, S.~Rane, J.~S. Yedidia, S.~C. Draper, and A.~Vetro, ``Feature transformation of biometric templates for secure biometric systems based on error correcting codes,'' in \emph{{IEEE} Conf. Comput. Vis. Pattern Recognit. Workshops ({CVPRW})}, 2008, pp. 1--6.

\bibitem[Scheirer and Boult(2007)]{ScheirerBoult-CrackingFuzzyVaultBioEncryption-BSYM-2007}
W.~Scheirer and T.~Boult, ``Cracking fuzzy vault and biometric encryption,'' in \emph{Biometric Symposium}.\hskip 1em plus 0.5em minus 0.4em\relax {IEEE} Press, September 2007, pp. 1--6.

\bibitem[Kholmatov and Yanikoglu(2008)]{Kholmatov-CorrelationAttackFuzzyVault-SPIE-2008}
A.~Kholmatov and B.~Yanikoglu, ``Realization of correlation attack against the fuzzy vault scheme,'' in \emph{Security, forensics, steganography, and watermarking of multimedia contents X}, vol. 6819.\hskip 1em plus 0.5em minus 0.4em\relax SPIE, 2008, pp. 263--269.

\bibitem[Dodis et~al.(2004)Dodis, Reyzin, and Smith]{Dodi-FuzzyExtractor-ec2004}
Y.~Dodis, L.~Reyzin, and A.~Smith, ``Fuzzy extrators: How to generate strong secret keys from biometrics and other noisy data,'' in \emph{In Advances in cryptology - Eurocrypt04}.\hskip 1em plus 0.5em minus 0.4em\relax Springer, 2004, pp. 523--540.

\bibitem[Tams et~al.(2015)Tams, Mihăilescu, and Munk]{Tams-SecurityConsiderationsFuzzyVaults-2015}
B.~Tams, P.~Mihăilescu, and A.~Munk, ``Security considerations in minutiae-based fuzzy vaults,'' \emph{IEEE Transactions on Information Forensics and Security}, vol.~10, no.~5, pp. 985--998, 2015.

\bibitem[Lim et~al.(2015)Lim, Teoh, and Kim]{Lim-BiometricFeatureTypeTransformation-2015}
M.~Lim, A.~B.~J. Teoh, and J.~Kim, ``Biometric feature-type transformation: Making templates compatible for secret protection,'' \emph{IEEE Signal Processing Magazine}, vol.~32, no.~5, pp. 77--87, 2015.

\bibitem[Chang et~al.(2004)Chang, Zhang, and Chen]{Chang-BiometricsKeyGeneration-ICME-2004}
Y.~Chang, W.~Zhang, and T.~Chen, ``Biometrics-based cryptographic key generation,'' in \emph{{IEEE} Intl. Conf. Multimedia and Expo ({ICME})}, vol.~3, 2004, pp. 2203--2206.

\bibitem[Nagar et~al.(2012)Nagar, Nandakumar, and Jain]{Nagar-MultibiometricCryptosystemsFeauterFusion-TIFS-2012}
A.~Nagar, K.~Nandakumar, and A.~Jain, ``Multibiometric cryptosystems based on feature-level fusion,'' \emph{IEEE Transactions on Information Forensics and Security}, vol.~7, no.~1, pp. 255--268, 2012.

\bibitem[Merkle and Tams(2014)]{merkle-securityimprovedfuzzyvault-2014}
\BIBentryALTinterwordspacing
J.~Merkle and B.~Tams, ``Security of the improved fuzzy vault scheme in the presence of record multiplicity (full version),'' 2014. [Online]. Available: \url{https://arxiv.org/abs/1312.5225}
\BIBentrySTDinterwordspacing

\bibitem[Guruswami and Sudan(1998)]{Guruswami-ImprovedDecodingOfReedSolomon-1998}
V.~Guruswami and M.~Sudan, ``Improved decoding of reed-solomon and algebraic-geometric codes,'' in \emph{Proceedings 39th Annual Symposium on Foundations of Computer Science (Cat. No.98CB36280)}, 1998, pp. 28--37.

\bibitem[Rathgeb and Uhl(2011)]{Rathgeb-BTP-Survey-EURASIP-2011}
C.~Rathgeb and A.~Uhl, ``A survey on biometric cryptosystems and cancelable biometrics,'' \emph{{EURASIP} Journal on Information Security}, vol.~3, March 2011.

\bibitem[Kim et~al.(2023)Kim, Jain, and Liu]{kim-adafacequalityadaptivemargin-2023}
M.~Kim, A.~K. Jain, and X.~Liu, ``Adaface: Quality adaptive margin for face recognition,'' pp. 18\,750--18\,759, 2023.

\bibitem[Boutros et~al.(2022)Boutros, Kaehm, Fang, Kirchbuchner, Damer, and Kuijper]{Boutros-LowResolutionIris-BIOSIG-2022}
F.~Boutros, O.~Kaehm, M.~Fang, F.~Kirchbuchner, N.~Damer, and A.~Kuijper, ``Low-resolution iris recognition via knowledge transfer,'' in \emph{2022 International Conference of the Biometrics Special Interest Group (BIOSIG)}.\hskip 1em plus 0.5em minus 0.4em\relax IEEE, September 2022, pp. 1--5.

\bibitem[Osorio-Roig et~al.(2024)Osorio-Roig, González-Soler, Rathgeb, and Busch]{OsorioRoig-PrivacyPreservingMultiBiometricIndexing-TIFS-2024}
D.~Osorio-Roig, L.~J. González-Soler, C.~Rathgeb, and C.~Busch, ``Privacy-preserving multi-biometric indexing based on frequent binary patterns,'' in \emph{IEEE Transactions on Information Forensics and Security}, 2024, vol.~19, pp. 4835--4850.

\bibitem[Guo et~al.(2024)Guo, Ray, Izydorczak, Goldfeder, Lipson, and Xu]{Guo-UnveilingIntraPersonFingerprintSimilarity-2024}
G.~Guo, A.~Ray, M.~Izydorczak, J.~Goldfeder, H.~Lipson, and W.~Xu, ``Unveiling intra-person fingerprint similarity via deep contrastive learning,'' \emph{Science Advances}, vol.~10, no.~2, p. eadi0329, 2024.

\bibitem[Engelsma et~al.(2021)Engelsma, Cao, and Jain]{Engelsma-LearningFixedLengthFingerprintRepresentation-2019}
J.~J. Engelsma, K.~Cao, and A.~K. Jain, ``Learning a fixed-length fingerprint representation,'' \emph{IEEE Transactions on Pattern Analysis and Machine Intelligence}, vol.~43, no.~6, pp. 1981--1997, 2021.

\bibitem[Rohwedder et~al.(2023)Rohwedder, Osorio-Roig, Rathgeb, and Busch]{Rohwedder-FixedLengthFingerprintDNN-BIOSIG-2023}
T.~Rohwedder, D.~Osorio-Roig, C.~Rathgeb, and C.~Busch, ``Benchmarking fixed-length fingerprint representations across different embedding sizes and sensor types,'' in \emph{Proc. Intl. Conf. of the Biometrics Special Interest Group ({BIOSIG})}.\hskip 1em plus 0.5em minus 0.4em\relax Gesellschaft f{\"u}r Informatik e.V., September 2023.

\end{thebibliography}
\end{document}